\documentclass[final]{cvpr}

\usepackage{times}
\usepackage{epsfig}
\usepackage{graphicx}
\usepackage{amsmath}
\usepackage{amssymb}

\usepackage{subfigure}
\usepackage{comment}
\usepackage{color}
\usepackage{multirow}  
\usepackage[ruled,linesnumbered]{algorithm2e}

\usepackage{makecell}

\usepackage[pagebackref=true,breaklinks=true,colorlinks,bookmarks=false]{hyperref}

\def\cvprPaperID{5829} 
\def\confYear{CVPR 2021}

\begin{document}

\title{Cross-Iteration Batch Normalization}

\author{
Zhuliang~Yao\textsuperscript{\rm 1, 2 }\thanks{This work is done when Zhuliang Yao is an intern at Microsoft Research Asia. Correspondence to: Yue Cao (yuecao@microsoft.com).}
\qquad~Yue~Cao\textsuperscript{\rm 2}
\qquad~Shuxin~Zheng\textsuperscript{\rm 2}
\qquad~Gao~Huang\textsuperscript{\rm 1}
\qquad~Stephen~Lin\textsuperscript{\rm 2}\\
{\textsuperscript{\rm 1} Tsinghua University}
\qquad{\textsuperscript{\rm 2} Microsoft Research Asia} \\
\small{\texttt{\{yzl17@mails.,gaohuang@\}tsinghua.edu.cn}}
\qquad\small{\texttt{\{yuecao,Shuxin.Zheng,stevelin\}@microsoft.com}}
}

\maketitle

\begin{abstract}
A well-known issue of Batch Normalization is its significantly reduced effectiveness in the case of small mini-batch sizes. When a mini-batch contains few examples, the statistics upon which the normalization is defined cannot be reliably estimated from it during a training iteration. To address this problem, we present Cross-Iteration Batch Normalization (CBN), in which examples from multiple recent iterations are jointly utilized to enhance estimation quality. A challenge of computing statistics over multiple iterations is that the network activations from different iterations are not comparable to each other due to changes in network weights. We thus compensate for the network weight changes via a proposed technique based on Taylor polynomials, so that the statistics can be accurately estimated and batch normalization can be effectively applied. On object detection and image classification with small mini-batch sizes, CBN is found to outperform the original batch normalization and a direct calculation of statistics over previous iterations without the proposed compensation technique. Code is available at \url{https://github.com/Howal/Cross-iterationBatchNorm}.

\end{abstract}

\section{Introduction} \label{intro}

Batch Normalization (BN)~\cite{ioffe2015batch} has played a significant role in the success of deep neural networks. It was introduced to address the issue of internal covariate shift, where the distribution of network activations changes during training iterations due to the updates of network parameters. This shift is commonly believed to be disruptive to network training, and BN alleviates this problem through normalization of the network activations by their mean and variance, computed over the examples within the mini-batch at each iteration. With this normalization, network training can be performed at much higher learning rates and with less sensitivity to weight initialization.

In BN, it is assumed that the distribution statistics for the examples within each mini-batch reflect the statistics over the full training set. While this assumption is generally valid for large batch sizes, it breaks down in the \emph{small batch size regime} \cite{peng2018megdet,wu2018group,ioffe2017batch}, where noisy statistics computed from small sets of examples can lead to a dramatic drop in performance. This problem hinders the application of BN to memory-consuming tasks such as object detection \cite{ren2015faster,dai2017deformable}, semantic segmentation \cite{long2015fully,chen2017deeplab} and action recognition \cite{wang2018non}, where batch sizes are limited due to memory constraints.

Towards improving estimation of statistics in the small batch size regime, alternative normalizers have been proposed. Several of them, including Layer Normalization (LN)~\cite{ba2016layer}, Instance Normalization (IN)~\cite{ulyanov2016instance}, and Group Normalization (GN)~\cite{wu2018group}, compute the mean and variance over the channel dimension, independent of batch size. Different channel-wise normalization techniques, however, tend to be suitable for different tasks, depending on the set of channels involved. 
Although GN is designed for detection task, the slow inference speed limits its practical usage. 
On the other hand, synchronized BN (SyncBN)~\cite{peng2018megdet} yields consistent improvements by processing larger batch sizes across multiple GPUs. These gains in performance come at the cost of additional overhead needed for synchronization across the devices.

\begin{figure}
\centering
    \includegraphics[width=0.6\columnwidth]{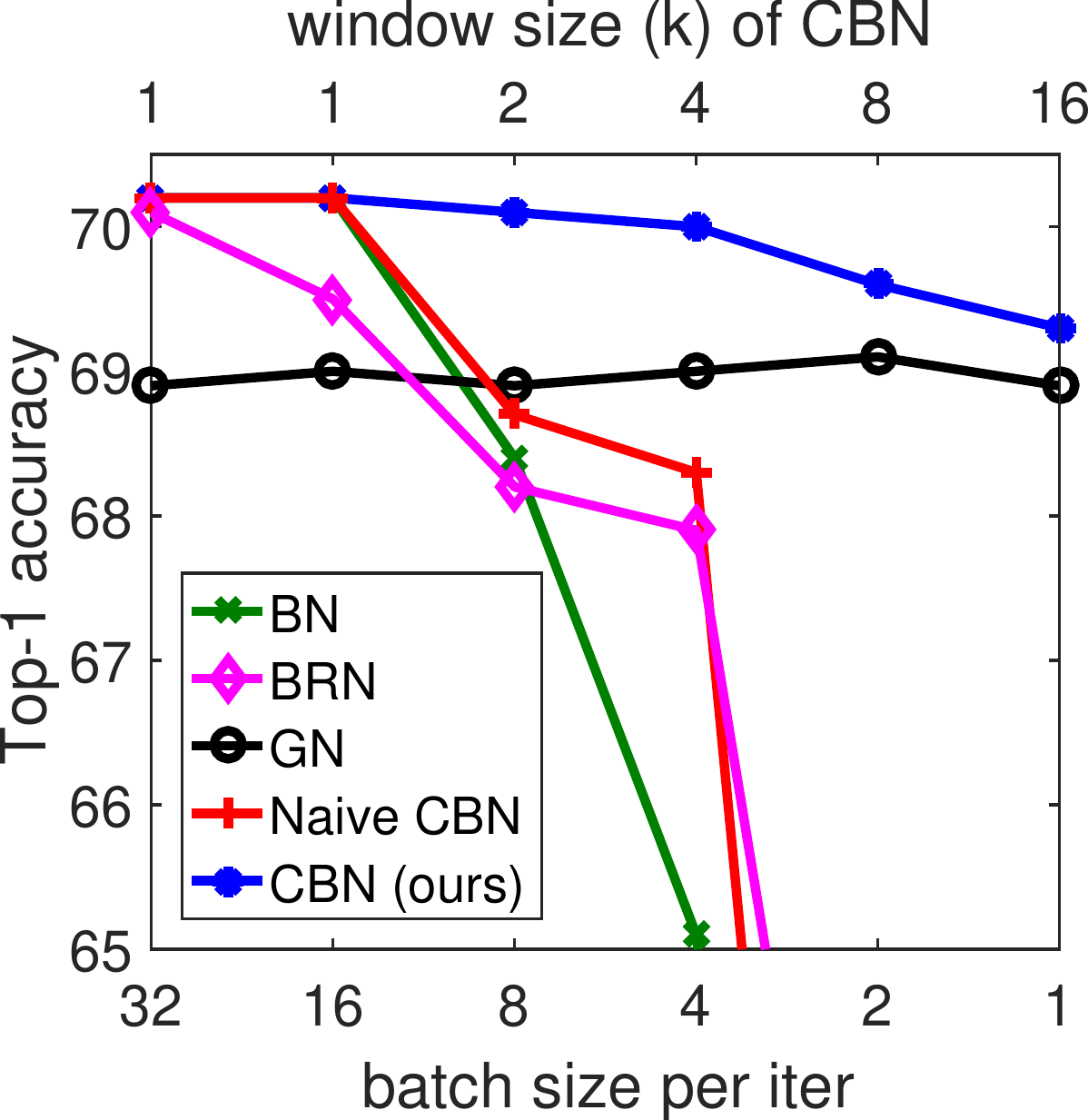}
    \caption{\textbf{Top-1 classification accuracy \emph{vs.} batch sizes per iteration.} The base model is a ResNet-18~\cite{he2016deep} trained on ImageNet~\cite{russakovsky2015imagenet}. The accuracy of BN~\cite{ioffe2015batch} drops rapidly when the batch size is reduced. BRN~\cite{ioffe2017batch} stabilizes BN a little but still has trouble with smaller batch sizes. GN~\cite{wu2018group} exhibits stable performance but underperforms BN on adequate batch sizes. CBN compensates for the reduced batch size per GPU by exploiting approximated statistics from recent iterations (Temporal window size denotes how many recent iterationss are utilized for statistics computation). CBN shows relatively stable performance over different batch sizes. Naive CBN, which directly calculates statistics from recent iterations without compensation, is shown not to work well.}\label{fig:teaser}
\end{figure}

A seldom explored direction for estimating better statistics is to compute them over the examples from multiple recent training iterations, instead of from only the current iteration as done in previous techniques. This can substantially enlarge the pool of data from which the mean and variance are obtained. However, there exists an obvious drawback to this approach, in that the activation values from different iterations are not comparable to each other due to the changes in network weights. As shown in Figure~\ref{fig:teaser}, directly calculating the statistics over multiple iterations, which we refer to as Naive CBN, results in lower accuracy.

In this paper, we present a method that compensates for the network weight changes among iterations, so that examples from preceding iterations can be effectively used to improve batch normalization. Our method, called Cross-Iteration Batch Normalization (CBN), is motivated by the observation that network weights change gradually, instead of abruptly, between consecutive training iterations, thanks to the iterative nature of Stochastic Gradient Descent (SGD). As a result, the mean and variance of examples from recent iterations can be well approximated for the current network weights via a low-order Taylor polynomial, defined on gradients of the statistics with respect to the network weights. The compensated means and variances from multiple recent iterations are averaged with those of the current iteration to produce better estimates of the statistics.

In the small batch size regime, CBN leads to appreciable performance improvements over the original BN, as exhibited in Figure~\ref{fig:teaser}. The superiority of our proposed approach is further demonstrated through more extensive experiments on ImageNet classification and object detection on COCO. These gains are obtained with negligible overhead, as the statistics from previous iterations have already been computed and Taylor polynomials are simple to evaluate. With this work, it is shown that cues for batch normalization can successfully be extracted along the time dimension, opening a new direction for investigation.

\section{Related Work}

The importance of normalization in training neural networks has been recognized for decades~\cite{lecun1998efficient}.
In general, normalization can be performed on three components: input data, hidden activations, and network parameters. Among them, input data normalization is used most commonly because of its simplicity and effectiveness~\cite{sola1997importance, lecun1998efficient}.

After the introduction of Batch Normalization~\cite{ioffe2015batch}, the normalization of activations has become nearly as prevalent. By normalizing hidden activations by their statistics within each mini-batch, BN effectively alleviates the vanishing gradient problem and significantly speeds up the training of deep networks. 
To mitigate the mini-batch size dependency of BN, a number of variants have been proposed, including Layer Normalization (LN)~\cite{ba2016layer}, Instance Normalization (IN)~\cite{ulyanov2016instance}, Group Normalization (GN)~\cite{wu2018group}, and Batch Instance Normalization (BIN)~\cite{nam2018batch}. The motivation of LN is to explore more suitable statistics for sequential models, while IN performs normalization in a manner similar to BN but with statistics only for each instance.
GN achieves a balance between IN and LN, by dividing features into multiple groups along the channel dimension and computing the mean and variance within each group for normalization. BIN introduces a learnable method for automatically switching between normalizing and maintaining style information, enjoying the advantages of both BN and IN on style transfer tasks.
Cross-GPU Batch Normalization (CGBN or SyncBN)~\cite{peng2018megdet} extends BN across multiple GPUs for the purpose of increasing the effective batch size. Though providing higher accuracy, it introduces synchronization overhead to the training process.
Kalman Normalization (KN)~\cite{wang2018kalman} presents a Kalman filtering procedure for estimating the statistics for a network layer from the layer's observed statistics and the computed statistics of previous layers.

Batch Renormalization (BRN)~\cite{ioffe2017batch} is the first attempt to utilize the statistics of recent iterations for normalization. It does not compensate for the statistics from recent iterations, but rather it down-weights the importance of statistics from distant iterations. This down-weighting heuristic, however, does not make the resulting statistics ``correct", as the statistics from recent iterations are not of the current network weights. BRN can be deemed as a special version of our Naive CBN baseline (without Taylor polynomial approximation), where distant iterations are down-weighted.

Recent work have also investigated the normalization of network parameters.
In Weight Normalization (WN)~\cite{salimans2016weight}, the optimization of network weights is improved through a reparameterization of weight vectors into their length and direction.
Weight Standardization (WS)~\cite{siyuan2019weight} instead reparameterizes weights based on their first and second moments for the purpose of smoothing the loss landscape of the optimization problem.
To combine the advantages of multiple normalization techniques, Switchable Normalization (SN)~\cite{luo2018differentiable} and Sparse Switchable Normalization (SSN)~\cite{shao2019ssn} make use of differentiable learning to switch among different normalization methods.

The proposed CBN takes an activation normalization approach that aims to mitigate the mini-batch dependency of BN. Different from existing techniques, it provides a way to effectively aggregate statistics across multiple training iterations.

\section{Method}  \label{sec:method}

\begin{figure*}[t]
\centering
\includegraphics[width=0.8\linewidth]{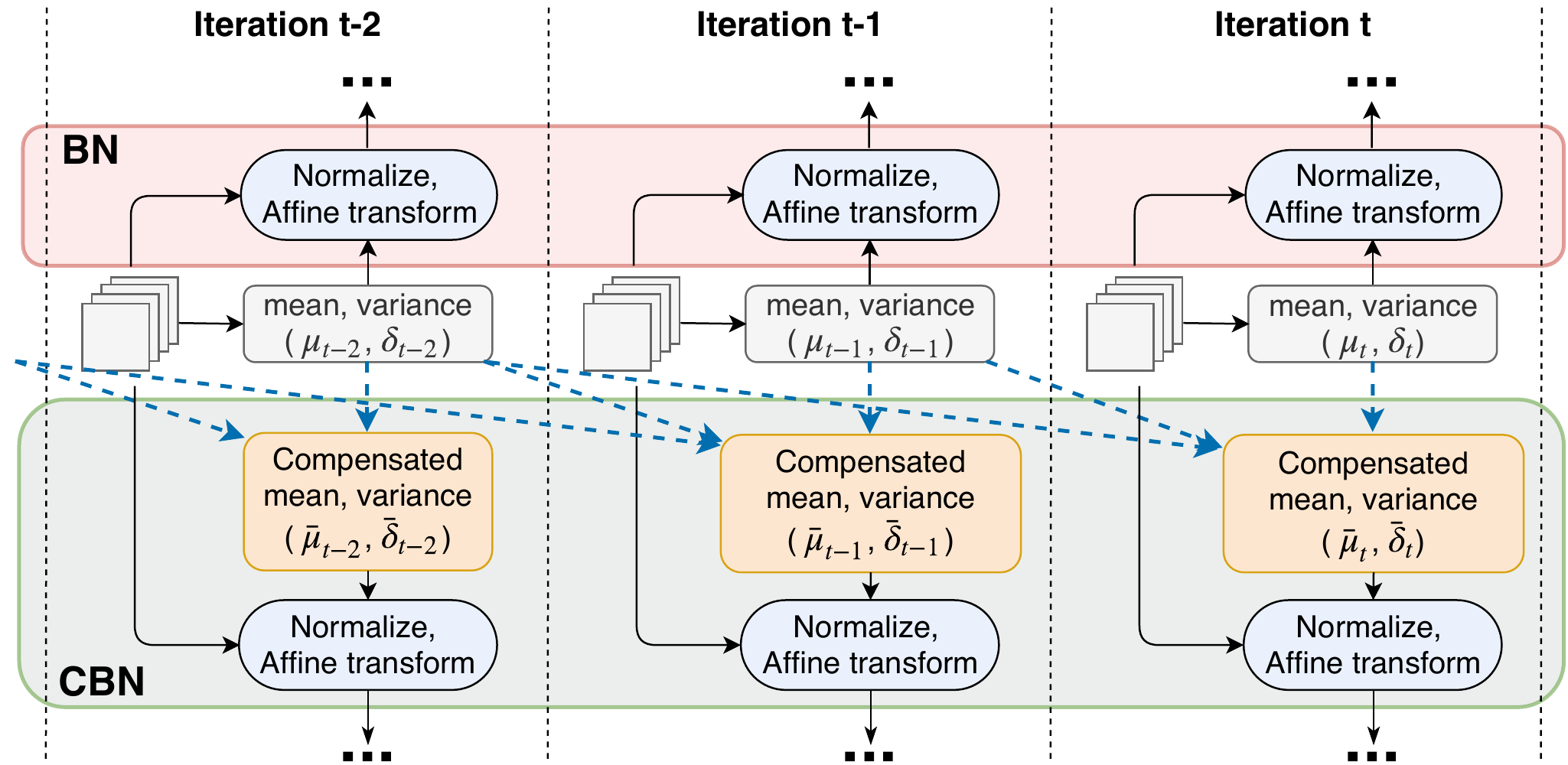}
\caption{Illustration of BN and the proposed Cross-Iteration Batch Normalization (CBN).}
\label{fig:asyncbn}
\vspace{-10pt}
\end{figure*}

\subsection{Revisiting Batch Normalization}

The original batch normalization (BN)~\cite{ioffe2015batch} whitens the activations of each layer by the statistics computed within a mini-batch. Denote $\theta_t$ and $x_{t,i}(\theta_t)$ as the network weights and the feature response of a certain layer for the $i$-th example in the $t$-th mini-batch. With these values, BN conducts the following normalization:
\begin{equation}
\hat{x}_{t,i}(\theta_t)=\frac{x_{t,i}(\theta_t)-\mu_t(\theta_t)}{\sqrt{\sigma_t(\theta_t)^2 + \epsilon}},
\end{equation}
where $\hat{x}_{t,i}(\theta_t)$ is the whitened activation with zero mean and unit variance, $\epsilon$ is a small constant added for numerical stability, and $\mu_t(\theta_t)$ and $\sigma_t(\theta_t)$ are the mean and variance computed for all the examples from the current mini-batch, i.e.,
\begin{equation}
    \mu_{t}(\theta_t) = \frac{1}{m} \sum_{i=1}^m x_{t,i}(\theta_t),\label{eq:bn_mu}
\end{equation}
\begin{equation}
\begin{split}
    \sigma_{t}(\theta_t) &= \sqrt{\frac{1}{m} \sum_{i=1}^m (x_{t,i}(\theta_t)-\mu_t(\theta_t))^2}\\
    &=\sqrt{\nu_t(\theta_t) - \mu_t(\theta_t)^2},\label{eq:bn_sigma}
\end{split}
\end{equation}
where $\nu_{t}(\theta_t)=\frac{1}{m} \sum_{i=1}^m x_{t,i}(\theta_t)^2$, and $m$ denotes the number of examples in the current mini-batch. The whitened activation $\hat{x}_{t,i}(\theta_t)$ further undergoes a linear transform with learnable weights, to increase its expressive power:
\begin{align}
y_{t,i}(\theta_t) = \gamma \hat{x}_{t,i}(\theta_t) + \beta,
\end{align}
where $\gamma$ and $\beta$ are the learnable parameters (initialized to $\gamma=1$ and $\beta=0$ in this work).

When the batch size $m$ is small, the statistics $\mu_{t}(\theta_t)$ and $\sigma_{t}(\theta_t)$ become noisy estimates of the training set statistics, thus degrading the effects of batch normalization. In the ImageNet classification task for which the BN module was originally designed, a batch size of 32 is typical. However, for other tasks requiring larger models and/or higher image resolution, such as object detection, semantic segmentation and video recognition, the typical batch size may be as small as 1 or 2 due to GPU memory limitations. The original BN becomes considerably less effective in such cases.

\subsection{Leveraging Statistics from Previous Iterations}

To address the issue of BN with small mini-batches, a naive approach is to compute the mean and variance over the current and previous iterations. However, the statistics $\mu_{t-\tau}(\theta_{t-\tau})$ and $\nu_{t-\tau}(\theta_{t-\tau})$ of the $(t-\tau)$-th iteration are computed under the network weights $\theta_{t-\tau}$, making them obsolete for the current iteration. As a consequence, directly aggregating statistics from multiple iterations produces inaccurate estimates of the mean and variance, leading to significantly worse performance.

We observe that the network weights change smoothly between consecutive iterations, due to the nature of gradient-based training. This allows us to approximate $\mu_{t-\tau}(\theta_{t})$ and $\nu_{t-\tau}(\theta_{t})$ from the readily available $\mu_{t-\tau}(\theta_{t-\tau})$ and $\nu_{t-\tau}(\theta_{t-\tau})$ via a Taylor polynomial, i.e.,
\begin{equation}
\begin{split}
    \mu_{t-\tau}(\theta_{t}) = &\mu_{t-\tau}(\theta_{t-\tau}) + \frac{\partial \mu_{t-\tau}(\theta_{t-\tau})}{\partial \theta_{t-\tau}} (\theta_{t}-\theta_{t-\tau})\\
   & + \mathbf{O}(||\theta_{t}-\theta_{t-\tau}||^2),\label{eq:taylor_mu}
\end{split}
\end{equation}
\begin{equation}
\begin{split}
    \nu_{t-\tau}(\theta_{t}) = &\nu_{t-\tau}(\theta_{t-\tau}) + \frac{\partial \nu_{t-\tau}(\theta_{t-\tau})}{\partial \theta_{t-\tau}} (\theta_{t}-\theta_{t-\tau})\\
    &+\mathbf{O}(||\theta_{t}-\theta_{t-\tau}||^2)\label{eq:taylor_nu},
\end{split}
\end{equation}
where ${\partial \mu_{t-\tau}(\theta_{t-\tau})}/{\partial \theta_{t-\tau}}$ and ${\partial \nu_{t-\tau}(\theta_{t-\tau})}/{\partial \theta_{t-\tau}}$ are gradients of the statistics with respect to the network weights, and $\mathbf{O}(||\theta_{t}-\theta_{t-\tau}||^2)$ denotes higher-order terms of the Taylor polynomial, which can be omitted since the first-order term dominates when $(\theta_{t}-\theta_{t-\tau})$ is small.

In Eq.~\eqref{eq:taylor_mu} and Eq.~\eqref{eq:taylor_nu}, the gradients ${\partial \mu_{t-\tau}(\theta_{t-\tau})}/{\partial \theta_{t-\tau}}$ and ${\partial \nu_{t-\tau}(\theta_{t-\tau})}/{\partial \theta_{t-\tau}}$ cannot be precisely determined at a negligible cost because the statistics $\mu_{t-\tau}^l (\theta_{t-\tau})$ and $\nu_{t-\tau}^l (\theta_{t-\tau})$ for a node at the $l$-th network layer depend on all the network weights prior to the $l$-th layer, i.e., ${\partial \mu_{t-\tau}^l(\theta_{t-\tau})}/{\partial \theta^r_{t-\tau}} \ne 0$ and ${\partial \nu^l_{t-\tau}(\theta_{t-\tau})}/{\partial \theta^r_{t-\tau}} \ne 0$ for $r \le l$, where $\theta^r_{t-\tau}$ denotes the network weights at the $r$-th layer. 
Only when $r=l$ can these gradients be derived in closed form efficiently.

Empirically, we find that as the layer index $r$ decreases ($r \leq l$), the partial gradients $\frac{\partial \mu_t^l(\theta_t)}{\theta_t^r}$ and $\frac{\partial \nu_t^l(\theta_t)}{\theta_t^r}$ rapidly diminish. 
These reduced effects of network weight changes at earlier layers on the activation distributions in later layers may perhaps be explained by the reduced internal covariate shift of BN.
Motivated by this phenomenon, which is studied in Appendix \ref{sec:grad-exp}, we propose to truncate these partial gradients at layer $l$.

Thus, we further approximate Eq.~\eqref{eq:taylor_mu} and Eq.~\eqref{eq:taylor_nu} by
\begin{small}
\begin{align}
    \mu_{t-\tau}^l (\theta_{t}) \approx \mu_{t-\tau}^l (\theta_{t-\tau}) + \frac{\partial \mu_{t-\tau}^l (\theta_{t-\tau})}{\partial \theta_{t-\tau}^l } (\theta_{t}^l -\theta_{t-\tau}^l),\label{eq:taylor_mu_layer}\\
    \nu_{t-\tau}^l (\theta_{t}) \approx \nu_{t-\tau}^l (\theta_{t-\tau}) + \frac{\partial \nu_{t-\tau}^l (\theta_{t-\tau})}{\partial \theta_{t-\tau}^l } (\theta_{t}^l -\theta_{t-\tau}^l).\label{eq:taylor_nu_layer}
\end{align}
\end{small}
A naive implementation of ${\partial \mu_{t-\tau}^l (\theta_{t-\tau})}/{\partial \theta_{t-\tau}^l }$ and ${\partial \nu_{t-\tau}^l (\theta_{t-\tau})}/{\partial \theta_{t-\tau}^l }$ involves computational overhead of $O(C_{out}\times C_{out} \times C_{in} \times K)$, where $C_{out}$ and $C_{in}$ denote the output and input channel dimension of the $l$-th layer, respectively, and $K$ denotes the kernel size of $\theta_{t-\tau}^l$. Here, we find that the operation can be implemented efficiently in $O(C_{out} \times C_{in} \times K)$, thanks to the averaging over feature responses of $\mu$ and $\nu$. See Appendix \ref{appendix:efficient-imp} for the details.

\subsection{Cross-Iteration Batch Normalization}

After compensating for network weight changes, we aggregate the statistics of the $k-1$ most recent iterations with those of the current iteration $t$ to obtain the statistics used in CBN:
\begin{align}
    \bar{\mu}^l_{t,k}(\theta_t) &= \frac{1}{k} \sum_{\tau=0}^{k-1} \mu_{t-\tau}^l(\theta_t),\label{eq:average_mu}\\
    \bar{\nu}^l_{t,k}(\theta_t) &= \frac{1}{k} \sum_{\tau=0}^{k-1} \max\big[ \nu_{t-\tau}^l(\theta_t), \mu_{t-\tau}^l(\theta_t)^2 \big ], \label{eq:average_nu_valid}
\end{align}
\begin{align}
    \bar{\sigma}^l_{t,k}(\theta_t) &= \sqrt{\bar{\nu}^l_{t,k}(\theta_t) - \bar{\mu}^l_{t,k}(\theta_t)^2},\label{eq:average_sigma}
\end{align}
where $\mu_{t-\tau}^l(\theta_t)$ and $\nu_{t-\tau}^l(\theta_t)$ are computed from Eq.~\eqref{eq:taylor_mu_layer} and Eq.~\eqref{eq:taylor_nu_layer}. In Eq.~\eqref{eq:average_nu_valid}, $\bar{\nu}^l_{t,k}(\theta_t)$ is determined from the maximum of $\nu_{t-\tau}^l(\theta_t)$ and $\mu_{t-\tau}^l(\theta_t)^2$ in each iteration because $\nu_{t-\tau}^l(\theta_t)\ge\mu_{t-\tau}^l(\theta_t)^2$ should hold for valid statistics but may be violated by Taylor polynomial approximations in Eq.~\eqref{eq:taylor_mu_layer} and Eq.~\eqref{eq:taylor_nu_layer}. Finally, $\bar{\mu}^l_{t,k}(\theta_t)$ and $\bar{\sigma}^l_{t,k}(\theta_t)$ are applied to normalize the corresponding feature responses $\{x^l_{t,i}(\theta_t)\}_{i=1}^m$ at the current iteration:
\begin{equation}
\hat{x}^l_{t,i}(\theta_t)=\frac{x^l_{t,i}(\theta_t)-\bar{\mu}^l_{t,k}(\theta_t)}{\sqrt{\bar{\sigma}^l_{t,k}(\theta_t)^2 + \epsilon}}.
\end{equation}

With CBN, the effective number of examples used to compute the statistics for the current iteration is $k$ times as large as that for the original BN. In training, the loss gradients are backpropagated to the network weights and activations at the current iteration, i.e., $\theta^l_t$ and $x^l_{t,i}(\theta_t)$. Those of the previous iterations are fixed and do not receive gradients.
Hence, the computation cost of CBN in back-propagation is the same as that of BN.

\begin{table*}[t]
\small
\centering
\addtolength{\tabcolsep}{0.0pt}
\begin{tabular}{c|ccc}
\Xhline{1.0pt}
 & batch size per iter & \#examples for statistics & Norm axis \\
\hline
IN  & \#bs/GPU * \#GPU & 1 & (spatial)  \\
LN  & \#bs/GPU * \#GPU & 1 & (channel, spatial)  \\
GN  & \#bs/GPU * \#GPU & 1 &  (channel group, spatial) \\
BN  & \#bs/GPU * \#GPU & \#bs/GPU & (batch, spatial) \\
syncBN  & \#bs/GPU * \#GPU & \#bs/GPU * \#GPU & (batch, spatial, GPU) \\
\hline
CBN  & \#bs/GPU * \#GPU & \#bs/GPU * \textbf{temporal window} & (batch, spatial, \textbf{iteration}) \\
\Xhline{1.0pt}
\end{tabular}
\caption{Comparison of different feature normalization methods. \#bs/GPU denotes batch size per GPU.}
\label{table:notations}
\end{table*}

Replacing the BN modules in a network by CBN leads to only minor increases in computational overhead and memory footprint. For computation, the additional overhead mainly comes from computing the partial derivatives ${\partial \mu_{t-\tau}(\theta_{t-\tau})}/{\partial \theta^l_{t-\tau}}$ and ${\partial \nu_{t-\tau}(\theta_{t-\tau})}/{\partial \theta^l_{t-\tau}}$, which is insignificant in relation to the overhead of the whole network. For memory, the module requires access to the statistics ($\{\mu^l_{t-\tau}(\theta_{t-\tau})\}_{\tau=1}^{k-1}$ and $\{\nu^l_{t-\tau}(\theta_{t-\tau})\}_{\tau=1}^{k-1}$) and the gradients (${\{\partial \mu_{t-\tau}(\theta_{t-\tau})}/{\partial \theta^l_{t-\tau}}\}_{\tau=1}^{k-1}$ and $\{{\partial \nu_{t-\tau}(\theta_{t-\tau})}/{\partial \theta^l_{t-\tau}}\}_{\tau=1}^{k-1}$) computed for the most recent $k-1$ iterations, which is also minor compared to the rest of the memory consumed in processing the input examples. The additional computation and memory of CBN is reported for our experiments in Table \ref{table:flops-memory}.

A key hyper-parameter in the proposed CBN is the temporal window size, $k$, of recent iterations used for statistics estimation. A broader window enlarges the set of examples, but the example quality becomes increasingly lower for more distant iterations, since the differences in network parameters $\theta_{t}$ and $\theta_{t-\tau}$ become more significant and are compensated less well using a low-order Taylor polynomial. Empirically, we found that CBN is effective with a window size up to $k=8$ in a variety of settings and tasks. The only trick is that the window size should be kept small at the beginning of training, when the network weights change quickly. Thus, we introduce a burn-in period of length $T_{\text{burn-in}}$ for the window size, where $k=1$ and CBN degenerates to the original BN. In our experiments, the burn-in period is set to 25 epochs on ImageNet image classification and 3 epochs on COCO object detection by default.

Table~\ref{table:notations} compares CBN with other feature normalization methods. The key difference among these approaches is the axis along which the statistics are counted and the features are normalized. The previous techniques are all designed to exploit examples from the same iteration. By contrast, CBN explores the aggregation of examples along the temporal dimension. As the data utilized by CBN lies in a direction orthogonal to that of previous methods, the proposed CBN could potentially be combined with other feature normalization approaches to further enhance statistics estimation in certain challenging applications.

\section{Experiments}

\subsection{Image Classification on ImageNet}

\textbf{Experimental settings.} ImageNet \cite{russakovsky2015imagenet} is a benchmark dataset for image classification, containing 1.28M training images and 50K validation images from 1000 classes. We follow the standard setting in \cite{he2015resnet} to train deep networks on the training set and report the single-crop top-1 accuracy on the validation set.
Our preprocessing and augmentation strategy strictly follows the GN baseline \cite{wu2018group}.
We use a weight decay of 0.0001 for all weight layers, including $\gamma$ and $\beta$. 
We train standard ResNet-18 for 100 epochs on 4 GPUs, and decrease the learning rate by the cosine decay strategy \cite{He2019bagoftricks}.
We perform the experiments for five trials, and report their mean and standard deviation (error bar). 
ResNet-18 with BN is our base model. To compare with other normalization methods, we directly replace BN with IN, LN, GN, BRN, and our proposed CBN.

\textbf{Comparison of feature normalization methods.}
In Table~\ref{table:norm-imagenet}, we compare the performance of each normalization method with a batch size, 32, sufficient for computing reliable statistics.
Under this setting, BN clearly yields the highest top-1 accuracy.
Similar to results found in previous works \cite{wu2018group}, the performance of IN and LN is significantly worse than that of BN.
GN works well on image classification but falls short of BN by 1.2\%.
Among all the methods, our CBN is the only one that is able to achieve accuracy comparable to BN, as it converges to the procedure of BN at larger batch sizes.

\begin{table}[t]
\centering
\small
\centering
\addtolength{\tabcolsep}{-3.7pt}
\begin{tabular}{c|ccccc}
\Xhline{1.0pt}
 & IN & LN & GN & CBN & BN \\
\hline
Top-1 acc & 64.4$\pm$0.2 & 67.9$\pm$0.2 & 68.9$\pm$0.1 & \textbf{70.2$\pm$0.1} & \textbf{70.2$\pm$0.1}\\
\Xhline{1.0pt}
\end{tabular}
\caption{Top-1 accuracy of \textbf{feature normalization methods} using ResNet-18 on ImageNet.}
\label{table:norm-imagenet}
\vspace{-10pt}
\end{table}

\begin{table}[t]
\addtolength{\tabcolsep}{-1.5pt}
\small
\centering
\begin{tabular}{c|cccccc}
\Xhline{1.0pt}
batch size per GPU & 32 & 16 & 8 & 4 & 2 & 1\\
\hline
BN   & \textbf{70.2} & \textbf{70.2} & 68.4 & 65.1 & 55.9 & - \\
GN   & 68.9 & 69.0 & 68.9 & 69.0 & 69.1 & 68.9\\
BRN  & 70.1 & 69.5 & 68.2 & 67.9 & 60.3 & -\\
CBN  & \textbf{70.2} & \textbf{70.2} & \textbf{70.1} & \textbf{70.0} & \textbf{69.6} & \textbf{69.3}\\
\Xhline{1.0pt}
\end{tabular}
\caption{Top-1 accuracy of {normalization methods} with \textbf{different batch sizes} using ResNet-18 as the base model on ImageNet.}
\label{table:batchsize-imagenet}
\end{table}

\textbf{Sensitivity to batch size.}
We compare the behavior of CBN, original BN~\cite{ioffe2015batch}, GN~\cite{wu2018group}, and BRN~\cite{ioffe2017batch} at the same number of images per GPU on ImageNet classification.
For CBN, the recent iterations are utilized so as to ensure that the number of effective examples is no fewer than 16.
For BRN, the settings strictly follow the original paper.
We adopt a learning rate of 0.1 for the batch size of 32, and linearly scale the learning rate by $N/32$ for a batch size of $N$.

The results are shown in Table \ref{table:batchsize-imagenet}.
For the original BN, its accuracy drops noticeably as the number of images per GPU is reduced from 32 to 2.
BRN suffers a significant performance drop as well.
GN maintains its accuracy by utilizing the channel dimension but not batch dimension. For CBN, its accuracy holds by exploiting the examples of recent iterations. Also, CBN outperforms GN by 0.9\% on average top-1 accuracy with different batch sizes. This is reasonable, because the statistics computation of CBN introduces uncertainty caused by the stochastic batch sampling like in BN, but this uncertainty is missing in GN which results in some loss of regularization ability.
For the extreme case that the number of images per GPU is 1, BN and BRN fails to produce results,
while CBN outperforms GN by 0.4\% on top-1 accuracy in this case. 

\begin{table}[t]
\small
\centering
\addtolength{\tabcolsep}{-0.5pt}
\begin{tabular}{c|ccccc}
\Xhline{1.0pt}
 & \begin{tabular}[c]{@{}c@{}}BN\\ bs=32\end{tabular} & \begin{tabular}[c]{@{}c@{}}BN\\ bs=4\end{tabular} & GN & BRN & CBN\\
\hline
ResNet-50 & 76.1 & 72.2 & 75.5 & 73.8 & 76.0\\
VGG-16 & 73.3 & 68.2 & 72.7 & 70.3 & 73.1\\
Inception-v3 & 77.5 & 72.9 & 76.8 & 75.1 & 77.2\\
DenseNet-121 & 74.7 & 72.6 & 74.2 & 74.0 & 74.6\\
MobileNet-v2 & 71.6 & 67.3 & 71.0 & 70.7 & 71.6\\
\Xhline{1.0pt}
\end{tabular}
\label{table:diff-arch-imagenet}
\caption{Top-1 accuracy of normalization methods with \textbf{different network architectures} on ImageNet.}
\vspace{-10pt}
\end{table}

\begin{table*}[t]
\small
\centering
\addtolength{\tabcolsep}{0pt}
\begin{tabular}{c|c|ccc|ccc}
\Xhline{1.0pt}
    backbone & box head & AP${^\text{bbox}}$ & AP$^\text{bbox}_{50}$ & AP$^\text{bbox}_{75}$&AP$^\text{bbox}_\text{S}$&AP$^\text{bbox}_\text{M}$&AP$^\text{bbox}_\text{L}$ \\
\hline
    fixed BN & - & 36.9 & 58.2 & 39.9 & 21.2 & 40.8 & 46.9\\
    fixed BN & BN & 36.3 & 57.3 & 39.2 & 20.8 & 39.7 & 47.3\\
    fixed BN & syncBN & 37.7 & 58.5 & 41.1 & 22.0 & 40.9 & 49.0\\
    fixed BN & GN & 37.8 & 59.0 & 40.8 & 22.3 & 41.2 & 48.4\\
    fixed BN & BRN & 37.4 & 58.1 & 40.3 & 22.0 & 40.7 & 48.3\\
    fixed BN & CBN & 37.7 & 59.0 & 40.7 & 22.1 & 40.9 & 48.8\\
\hline
    BN & BN & 35.5 & 56.4 & 38.7 & 19.7 & 38.8 & 47.3\\
    syncBN & syncBN & 37.9 & 58.5 & 41.1 & 21.7 & 41.5 & 49.7\\
    GN & GN & 37.8 & 59.1 & 40.9 & 22.4 & 41.2 & 49.0\\
    CBN & CBN & 37.7 & 58.9 & 40.6 & 22.0 & 41.4 & 48.9\\
\Xhline{1.0pt}
\end{tabular}
\caption{Results of \textbf{feature normalization methods} on Faster R-CNN with FPN and ResNet50 on COCO. {As the values of standard deviation of all methods are less than 0.1 on COCO, we ignore them here.}}
\label{table:ablation-norm-coco}
\vspace{-5pt}
\end{table*}

\begin{table*}[t]
\small
\centering
\addtolength{\tabcolsep}{0pt}
\begin{tabular}{ccc|ccc|ccc}
\Xhline{1.0pt}
    Backbone & method & norm & AP${^\text{bbox}}$ & AP$^\text{bbox}_{50}$ & AP$^\text{bbox}_{75}$&AP$^\text{bbox}_\text{S}$&AP$^\text{bbox}_\text{M}$&AP$^\text{bbox}_\text{L}$ \\
\hline
    \multirow{3}{*}{R50+FPN} & \multirow{3}{*}{Faster R-CNN} & GN & 37.8 & 59.0 & 40.8 & 22.3 & 41.2 & 48.4\\
    & & syncBN & 37.7 & 58.5 & 41.1 & 22.0 & 40.9 & 49.0\\
    & & CBN & 37.7 & 59.0 & 40.7 & 22.1 & 40.9 & 48.8\\
    \hline
    \multirow{3}{*}{R101+FPN} & \multirow{3}{*}{Faster R-CNN} & GN & 39.3 & 60.6 & 42.7 & 22.5 & 42.5 & 51.3\\
    & & syncBN & 39.2 & 59.8 & 43.0 & 22.2 & 42.9 & 51.6\\
    & & CBN & 39.2 & 60.0 & 42.6 & 22.3 & 42.6 & 51.1\\
\Xhline{1.0pt}
     &  &  &  AP${^\text{bbox}}$ & AP$^\text{bbox}_{50}$ & AP$^\text{bbox}_{75}$&AP$^\text{mask}$&AP$^\text{mask}_{50}$&AP$^\text{mask}_{75}$ \\
    \cline{4-9}
    \multirow{3}{*}{R50+FPN} & \multirow{3}{*}{Mask R-CNN} & GN & 38.6 & 59.8 & 41.9 & 35.0 & 56.7 & 37.3\\
    & & syncBN & 38.5 & 58.9 & 42.3 & 34.7 & 56.3 & 36.8\\
    & & CBN & 38.5 & 59.2 & 42.1 & 34.6 & 56.4  & 36.6 \\
    \hline
    \multirow{3}{*}{R101+FPN} & \multirow{3}{*}{Mask R-CNN} & GN & 40.3 & 61.2 & 44.2 & 36.6 & 58.5 & 39.2\\
    & & syncBN & 40.3 & 60.8 & 44.2 & 36.0 & 57.7 & 38.6\\
    & & CBN & 40.1 & 60.5 & 44.1 & 35.8 & 57.3 & 38.5\\
\Xhline{1.0pt}
\end{tabular}
\caption{Results with \textbf{stronger backbones} on COCO object detection and instance segmentation.}
\label{table:ablation-backbone-coco}
\vspace{-10pt}
\end{table*}

\textbf{Different network architectures.}
To verify the generalization ability of CBN, we also compared CBN to BN and GN using different network architectures. The results are shown in Table \ref{table:diff-arch-imagenet}. We choose five typres of architectures, i.e., ResNet-50 \cite{he2016deep}, VGG-16 \cite{Simonyan14very}, Inception-v3 \cite{szegedy2017inceptionv4}, DenseNet-121 \cite{huang2019convolutional}, and MobileNet-v2 \cite{sandler2018mobilenetv2}. This set of architectures represents the majority of modern network choices for computer vision tasks. BN (bs=32) is the ideal upper bound of this experiment. All the other three normalization methods are trained with a batch size of 4. BN (bs=4) clearly suffers from the limitations of small batch-size regime. Also, GN leads to a about a 0.5\% performance drop. Our CBN is the only one that obtains results comparable to BN with large batch size. These results demonstrate that our proposed CBN can be used in most modern convolutional neural networks.

\subsection{Object Detection and Instance Segmentation on COCO}
\textbf{Experimental settings.}
COCO~\cite{lin2014microsoft} is chosen as the benchmark for object detection and instance segmentation. Models are trained on the COCO 2017 train split with 118k images, and evaluated on the COCO 2017 validation split with 5k images. Following the standard protocol in~\cite{lin2014microsoft}, the object detection and instance segmentation accuracies are measured by the mean average precision (mAP) scores at different intersection-over-union (IoU) overlaps at the box and the mask levels, respectively. 

Following~\cite{wu2018group}, Faster R-CNN~\cite{ren2015faster} and Mask R-CNN~\cite{he2017mask} with FPN~\cite{he2017fpn} are chosen as the baselines for object detection and instance segmentation, respectively. For both, the 2fc box head is replaced by a 4conv1fc head for better use of the normalization mechanism~\cite{wu2018group}. The backbone networks are ImageNet pretrained ResNet-50 (default) or ResNet-101, with specific normalization. Finetuning is performed on the COCO train set for 12 epochs on 4 GPUs by SGD, where each GPU processes 4 images (default). Note that the mean and variance statistics in CBN are computed within each GPU. The learning rate is initialized to be $0.02*N/16$ for a batch size per GPU of $N$, and is decayed by a factor of 10 at the 9-th and the 11-th epochs. The weight decay and momentum parameters are set to 0.0001 and 0.9, respectively. We use the average over 5 trials for all results. As the values of standard deviation of all methods are less than 0.1 on COCO, they are ignored here.

As done in~\cite{wu2018group}, we experiment with two settings where the normalizers are activated only at the task-specific heads with frozen BN at the backbone (default), or the normalizers are activated at all the layers except for the early conv1 and conv2 stages in ResNet.

\textbf{Normalizers at backbone and task-specific heads.}
We further study the effect of different normalizers on the backbone network and task-specific heads for object detection on COCO. CBN, original BN, syncBN, and GN are included in the comparison.
For BRN, it is unclear \cite{luo2018differentiable} how to apply it in tasks like object detection. Directly replacing BN with BRN leads to 0.3\% performance drop on AP${^\text{bbox}}$ score.

Table \ref{table:ablation-norm-coco} presents the results.
When BN is frozen in the backbone and no normalizer is applied at the head, the AP${^\text{bbox}}$ score is 36.9\%.
When the original BN is applied at the head only and at both the backbone and the head, the accuracy drops to 36.3\% and 35.5\%, respectively.
For CBN, the accuracy is 37.7\% and 37.7\% at these two settings, respectively. Without any synchronization across GPUs, CBN can achieve performance on par with syncBN and GN, showing the superiority of the proposed approach.

\textbf{Instance segmentation and stronger backbones.}
Results of object detection (Faster R-CNN \cite{ren2015faster}) and instance segmentation (Mask R-CNN \cite{he2017mask}) with ResNet-50 and ResNet-101 are presented in Table \ref{table:ablation-backbone-coco}.
We can observe that our CBN achieves performance comparable to syncBN and GN with R50 and R101 as the backbone on both Faster R-CNN and Mask R-CNN, which demonstrates that CBN is robust and versatile to various deep models and tasks.

\subsection{Ablation Study}

\begin{figure}[t]
    \centering
    \subfigure{
        \includegraphics[width=0.4\columnwidth]{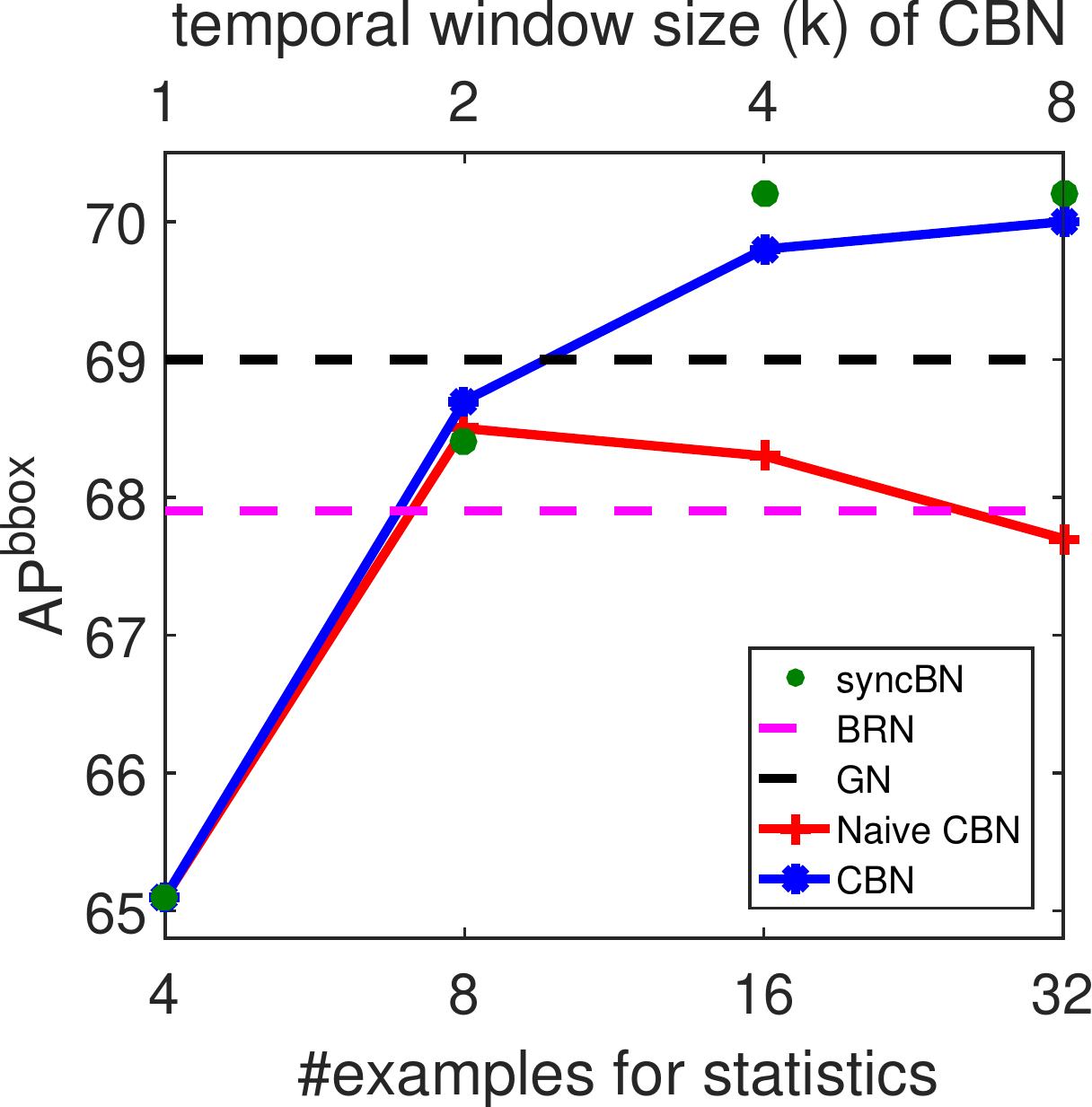}
        \label{fig:approx-batch-imagenet}
        }
    \subfigure{
        \includegraphics[width=0.43\columnwidth]{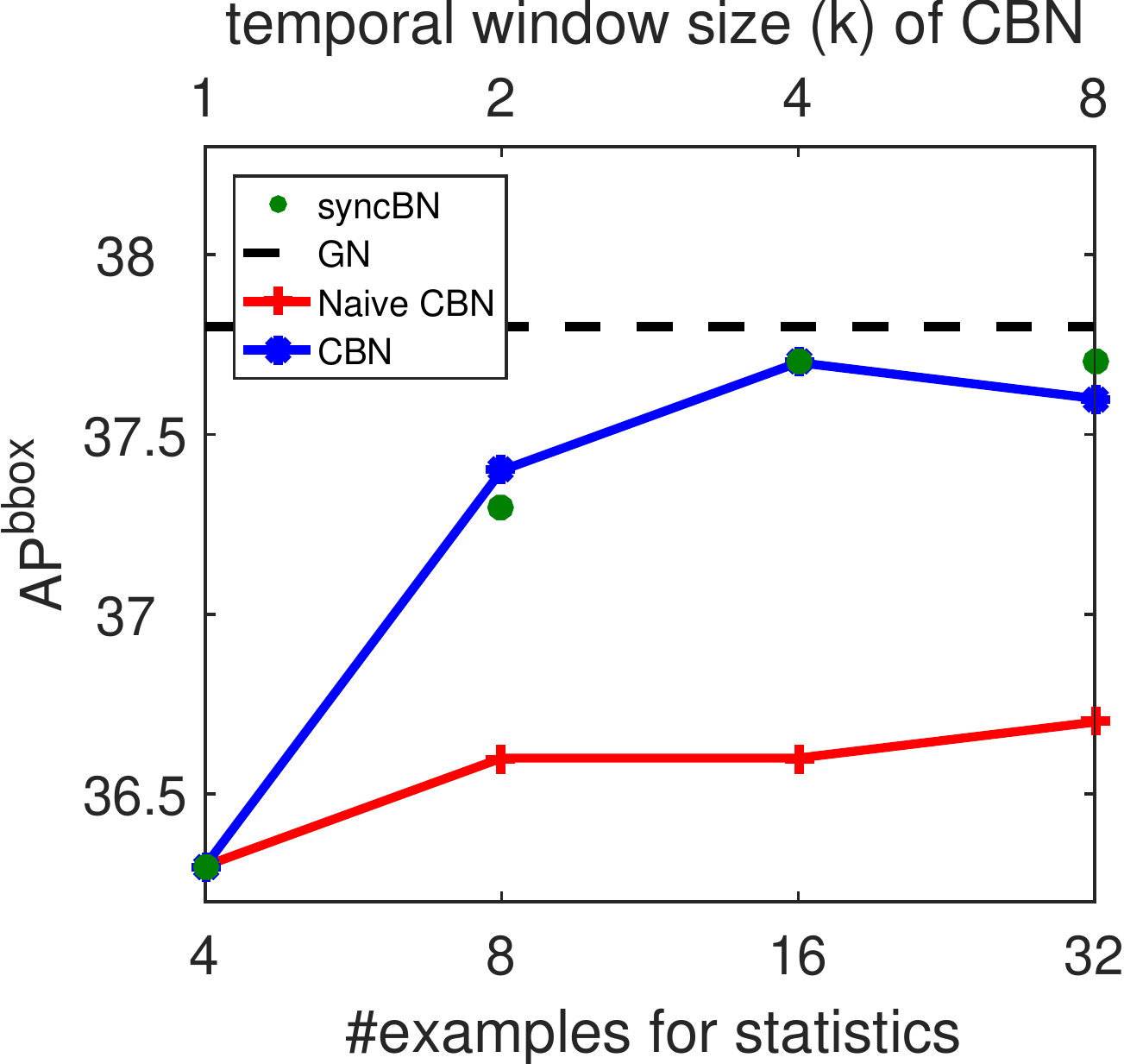}
        \label{fig:approx-batch-coco}
    }
    \caption{The effect of \textbf{temporal window size (k)} on ImageNet (ResNet-18) and COCO (Faster R-CNN with ResNet-50 and FPN) with \#bs/GPU = 4 for CBN and Naive CBN. Naive CBN directly utilizes statistics from recent iterations, while BN uses the equivalent \#examples as CBN for statistics computation.}
    \label{fig:approx-batch}
    \vspace{-10pt}
\end{figure}

\textbf{Effect of temporal window size $k$}.
We conduct this ablation on ImageNet image classification and COCO object detection, with each GPU processing 4 images.
Figure \ref{fig:approx-batch} presents the results. When $k=1$, only the batch from the current iteration is utilized; therefore, CBN degenerates to the original BN. The accuracy suffers due to the noisy statistics on small batch sizes.
As the window size $k$ increases, more examples from recent iterations are utilized for statistics estimation, leading to greater accuracy.
Accuracy saturates at $k=8$ and even drops slightly. For more distant iterations, the network weights differ more substantially and Taylor polynomial approximation becomes less accurate.

On the other hand, it is empirically observed that the original BN saturates at a batch size of 16 or 32 for numerous applications \cite{peng2018megdet,wu2018group}, indicating that the computed statistics become accurate. Thus, a temporal window size of $k = \min(\lceil \frac{16}{\text{bs per GPU}}\rceil, 8)$ is suggested.

\textbf{Effect of compensation.}
To study this, we compare CBN to 1) Naive CBN, where statistics from recent iterations are directly aggregated without compensation via Taylor polynomial; and 2) the original BN applied with the same effective example number as CBN (i.e., its batch size per GPU is set to the product of batch size per GPU and temporal window size of CBN), which does not require any compensation and serves as an upper performance bound.

The experimental results are also presented in Figure \ref{fig:approx-batch}.
CBN clearly surpasses Naive CBN when the previous iterations are included.
Actually, Naive CBN fails when the temporal window size grows to $k=8$ as shown in Figure \ref{fig:approx-batch-imagenet}, demonstrating the necessity of compensating for changing network weights over iterations.
Compared with the original BN upper bound, CBN achieves similar accuracy at the same effective example number. This result indicates that the compensation using a low-order Taylor polynomial by CBN is effective.

\begin{figure}[t]
    \centering
    \subfigure{
        \includegraphics[width=0.48\columnwidth]{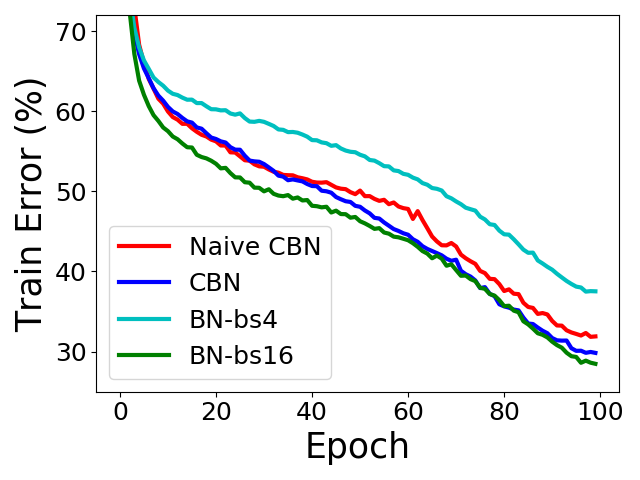}
        \label{fig:train-curves-cifar}
    }
    \subfigure{
        \includegraphics[width=0.48\columnwidth]{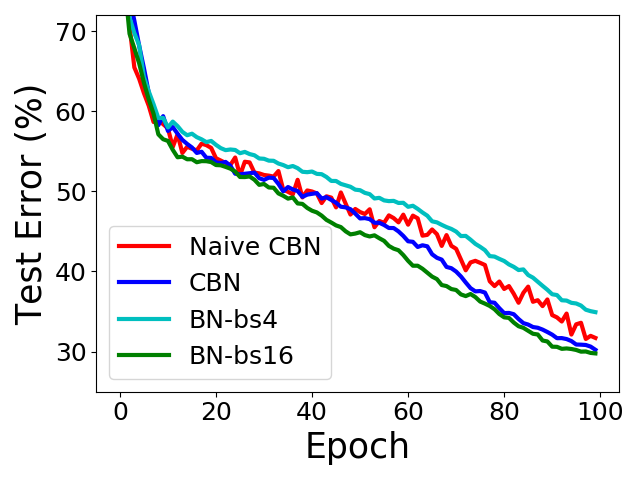}
        \label{fig:val-curves-cifar}
    }
    \vspace{-10pt}
    \caption{Training and test curves for CBN, Naive CBN, and BN on ImageNet, with batch size per GPU of 4 and temporal window size $k=4$ for CBN, Naive CBN, and BN-bs4, and batch size per GPU of 16 for BN-bs16. The plot of BN-bs16 is the ideal bound.}
    \label{fig:curves-cifar}
    \vspace{-5pt}
\end{figure}

Figure \ref{fig:curves-cifar} presents the train and test curves of CBN, Naive CBN, BN-bs4, and BN-bs16 on ImageNet, with 4 images per GPU and a temporal window size of 4 for CBN, Naive CBN, and BN-bs4, and 16 images per GPU for BN-bs16.
The train curve of CBN is close to BN-bs4 at the beginning, and approaches BN-bs16 at the end. The reason is that we adopt a burn-in period to avoid the disadvantage of rapid statistics change at the beginning of training.
The gap between the train curve of Naive CBN and CBN shows that Naive CBN cannot even converge well on the training set.
The test curve of CBN is close to BN-bs16 at the end, while Naive CBN exhibits considerable jitter.
All these phenomena indicate the effectiveness of our proposed Taylor polynomial compensation.

\begin{figure}[t]
    \centering
    \subfigure[ImageNet]{
        \includegraphics[width=0.45\columnwidth]{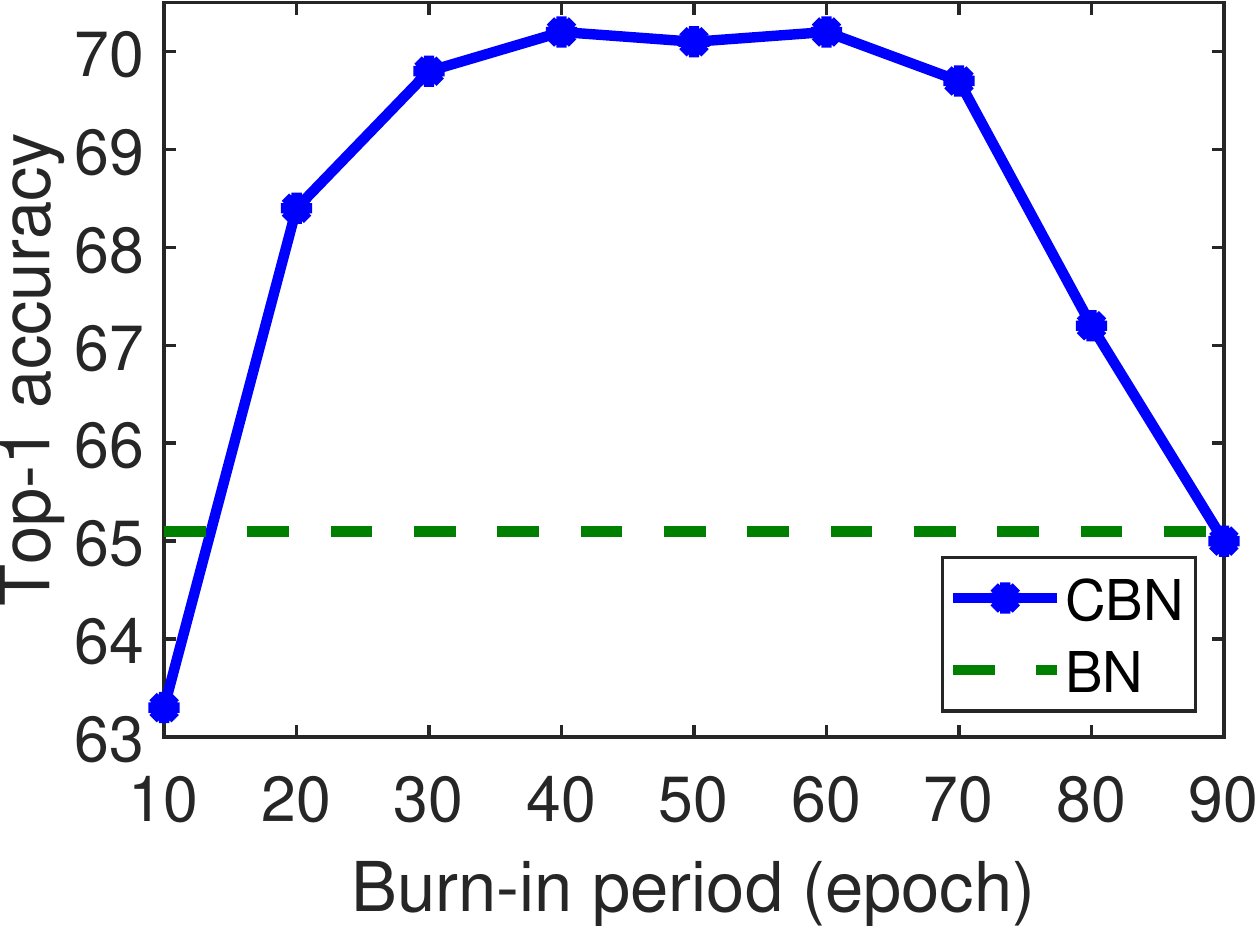}
        \label{fig:step-start-epoch-imagenet}
    }
    \subfigure[COCO]{
        \includegraphics[width=0.45\columnwidth]{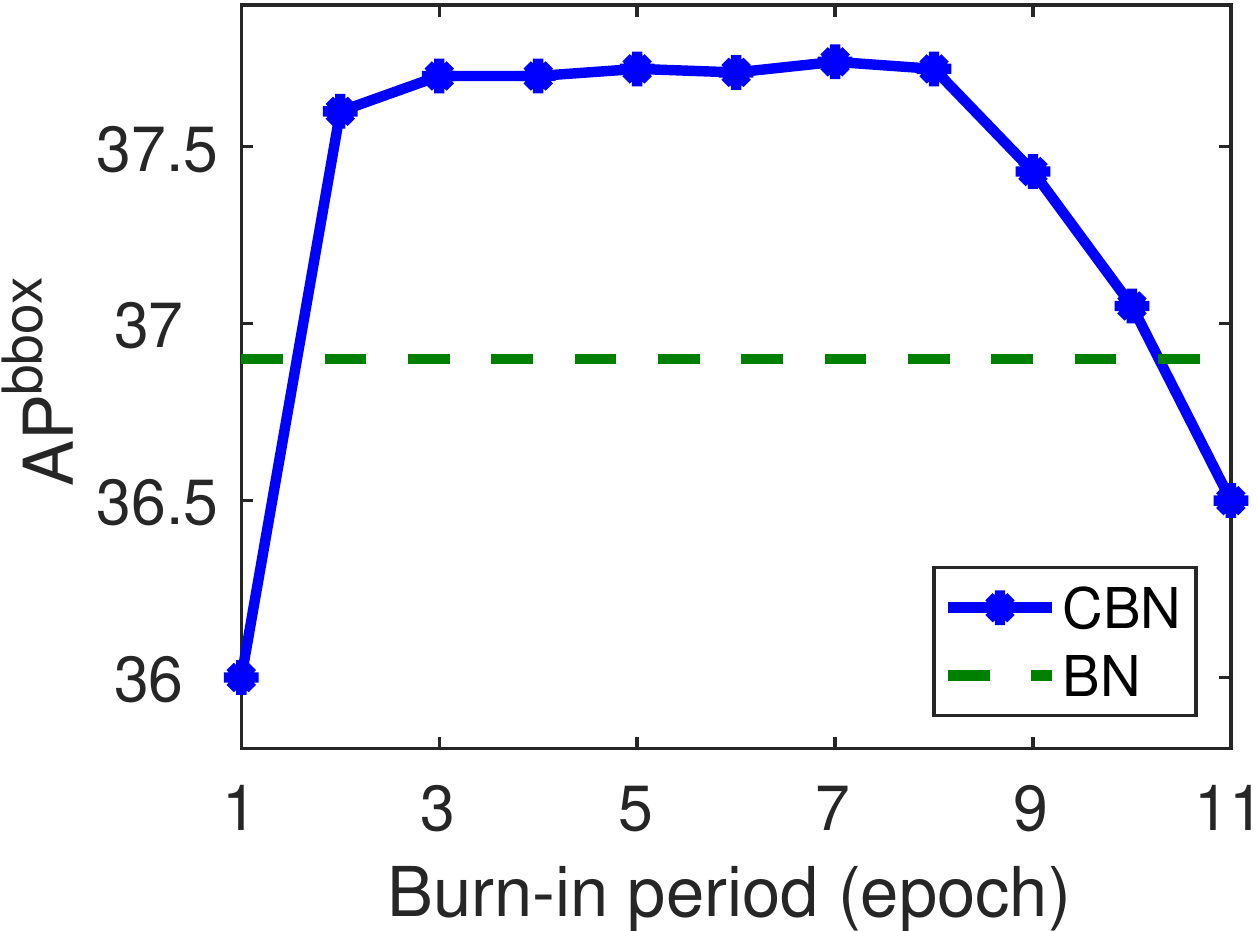}
        \label{fig:step-start-epoch-COCO}
    }
    \vspace{-3pt}
    \caption{Results of \textbf{different burn-in periods (in epochs)} on CBN, with batch size per iteration of 4, on ImageNet and COCO.}
    \label{fig:step-start-epoch}
    \vspace{-10pt}
\end{figure}

\textbf{Effect of burn-in period length $T$.}
We study the effect of varying the burn-in period length $T_{\text{burn-in}}$, at 4 images per GPU on both ImageNet image classification (ResNet-18) and COCO object detection (Faster R-CNN with FPN and ResNet-50).
Figure \ref{fig:step-start-epoch-imagenet} and \ref{fig:step-start-epoch-COCO} present the results. When the burn-in period is too short, the accuracy suffers. This is because at the beginning of training, the network weights change rapidly, causing the compensation across iterations to be less effective. When the burn-in period is too long, i.e., CBN is involved too late and the overall performance drops to the BN baseline.

\begin{table}[t]
\small
\centering
\addtolength{\tabcolsep}{-0.5pt}
\begin{tabular}{c|cccc}
\Xhline{1.0pt}
  & Epoch-8 & Epoch-9 & Epoch-10 & Epoch-11 \\ \hline
 AP${^\text{bbox}}$ & 37.7 & 37.7 & 37.6 & 37.3 \\ 
\Xhline{1.0pt}
\end{tabular}
\caption{Results on switching from BN to syncBN at different epochs on COCO.}
\label{table:ablation-switch-syncbn}
\vspace{-10pt}
\end{table}

An interesting observation is that the accuracy is stable for a wide range of burn-in periods $T_{\text{burn-in}}$. This leads to a question of whether BN in the small batch-size regime only suffers in terms of generalization performance in later stages of training.
For further exploration, we design an experiment to remove other influences: we first train the model on COCO with standard BN and a small batch size, then switch BN to syncBN. We present the experimental results in Table \ref{table:ablation-switch-syncbn}.
Results show that syncBN works similarly to CBN, which further verifies the high performance of CBN. It also supports our assumption that BN in the small batch-size regime only suffers in terms of generalization performance in later stages of training, which may shed some light on the small batch-size regime.

\subsection{Analysis}

\textbf{Computational cost, memory footprint, and training/inference speed.} We examine the computational cost, memory footprint, and the training and inference speed of BN, GN and CBN in a practical COCO object detection task using R50-Mask R-CNN, shown in Table \ref{table:flops-memory}. 
The batch size per GPU and window size of CBN are set to 4.

Compared to BN and GN, CBN consumes about 7\% extra memory and 11\% more computational cost. The extra memory mainly contains the statistics ($\mu$ and $\nu$), their respective gradients, and the network parameters ($\theta_{t-1}\cdots\theta_{t-(k-1)}$) of previous iterations, while the computational cost comes from calculations of the statistics' respective gradients, Taylor compensations, and averaging operations.

The overall training speed of CBN is close to both BN and GN.
It is worth noting that the inference speed of CBN is equal to BN, which is much faster than GN. The inference stage of CBN is the same as that of BN, where pre-recorded statistics can be used instead of online statistics calculation.
From these results, the additional overhead of CBN is seen to be minor.
Also, merging BN/CBN into convolution in inference~\cite{li2018deeprebirth} could be utilized for further speedup.

\begin{table}[t]
\small
\centering
\addtolength{\tabcolsep}{-2.5pt}
\begin{tabular}{c|c|c|c|c}
\Xhline{1.0pt}
 & \begin{tabular}[c]{@{}c@{}}Memory\\ (GB)\end{tabular} & \begin{tabular}[c]{@{}c@{}}FLOPs\\ (M)\end{tabular} & \begin{tabular}[c]{@{}c@{}}Training\\ Speed (iter/s)\end{tabular} & \begin{tabular}[c]{@{}c@{}}Inference\\ Speed (iter/s)\end{tabular} \\ \hline
BN & 14.1 & 5155.1 & 1.3 & 6.2 \\ \hline
GN & 14.1 & 5274.2 & 1.2 & 3.7 \\ \hline
CBN & 15.1 & 5809.7 & 1.0 & 6.2 \\ \Xhline{1.0pt}
\end{tabular}
\caption{Comparison of theoretical \textbf{memory, FLOPs} and practical \textbf{training and inference speed} between original BN, GN, and CBN in both training and inference on COCO.}
\label{table:flops-memory}
\vspace{-5pt}
\end{table}

\begin{table}[t]
\small
\centering
\addtolength{\tabcolsep}{-0.5pt}
\begin{tabular}{c|cccc}
\Xhline{1.0pt}
 & BN & Naive CBN & CBN$^{(1)}$ & CBN$^{(2)}$ \\ \hline
 Top-1 acc & 65.1 & 66.8 & 70.0 & 70.0\\ 
\Xhline{1.0pt}
\end{tabular}
\caption{Top-1 accuracy of CBN that compensating with \textbf{different orders} and batch size per iter = 4 on ImageNet.}
\label{table:ablation-order-imagenet}
\vspace{-10pt}
\end{table}

\textbf{Using second order statistics for compensation.}
Results of CBN with different orders of Taylor expansion (batch size = 4, \#iterations for approximation = 3) are shown in Table \ref{table:ablation-order-imagenet}. 
By directly using the statistics of recent iterations without compensation, Naive CBN outperforms BN with batch size 4 by 1.7\% in accuracy.
Via compensating the statistics of recent iterations with a first-order Taylor expansion, CBN$^{(1)}$ can further improve the accuracy by 3.2\% compared to Naive CBN.
However, CBN$^{(2)}$ using a second-order approximation does not achieve better performance than CBN$^{(1)}$.
This may be because CBN$^{(1)}$ already achieves performance comparable to BN with large batch size, which serves as the upper bound of our approach, indicating that a first-order approximation is enough for image classification on ImageNet. Therefore, first-order compensation for CBN is adopted by default.

\textbf{Using more than one layer for compensation.}
We also study the influence of applying compensation over more than one layer.
CBN using two layers for compensation achieves 70.1 on ImageNet (batch size per GPU=4, k=4), which is comparable to CBN using only one layer.
However, the efficient implementation can no longer be used when more than one layer of compensation is employed.
As using more layers does not further improve performance but consumes more FLOPs, we adopt one-layer compensation for CBN in practice.

\section{Conclusion}

In the small batch size regime, batch normalization is widely known to drop dramatically in performance. To address this issue, we propose to enhance the quality of statistics via utilizing examples from multiple recent iterations. As the network activations from different iterations are not comparable to each other due to changes in network weights, we compensate for the network weight changes based on Taylor polynomials, so that the statistics can be accurately estimated. In the experiments, the proposed approach is found to outperform original batch normalization and a direct calculation of statistics over previous iterations without compensation. Moreover, it achieves performance on par with SyncBN, which can be regarded as the upper bound, on both ImageNet and COCO object detection.

{\small
\bibliographystyle{ieee_fullname}
\bibliography{refs}
}

\clearpage
\appendix
\section{Algorithm Outline}
Algorithm~\ref{alg:cbn} presents an outline of our proposed Cross-Iteration Batch Normalization (CBN).

\begin{algorithm}[ht]
\caption{Cross-Iteration Batch Normalization(CBN)}
\label{alg:cbn}
\KwIn{Feature responses of a network node of the $l$-th layer at the $t$-th iteration $\{x^l_{t,i}(\theta_t)\}_{i=1}^m$, network weights $\{\theta^l_{t-\tau}\}_{\tau=0}^{k-1}$, statistics $\{\mu^l_{t-\tau}(\theta_{t-\tau})\}_{\tau=1}^{k-1}$ and $\{\nu^l_{t-\tau}(\theta_{t-\tau})\}_{\tau=1}^{k-1}$, and gradients ${\{\partial \mu_{t-\tau}(\theta_{t-\tau})}/{\partial \theta^l_{t-\tau}}\}_{\tau=1}^{k-1}$ and $\{{\partial \nu_{t-\tau}(\theta_{t-\tau})}/{\partial \theta^l_{t-\tau}}\}_{\tau=1}^{k-1}$ from most recent $k-1$ iterations}
\KwOut{$\{y^l_{t,i}(\theta_t)=\text{CBN}(x^l_{t,i}(\theta_t))\}$}
$\mu_t(\theta_t) \leftarrow \frac{1}{m}\sum_{i=1}^m x_{t,i}(\theta_t)$,
$\nu_t(\theta_t) \leftarrow \frac{1}{m}\sum_{i=1}^m x_{t,i}^2(\theta_t)$ \hfill //statistics on the current iteration\\
\For{$\tau \in \{1,\ldots, k\}$}{
$\mu_{t-\tau}^l (\theta_{t}) \leftarrow \mu_{t-\tau}^l (\theta_{t-\tau}) + \frac{\partial \mu_{t-\tau}^l (\theta_{t-\tau})}{\partial \theta_{t-\tau}^l } (\theta_{t}^l -\theta_{t-\tau}^l)$ \hfill //approximation from recent iterations \\
$\nu_{t-\tau}^l (\theta_{t}) \leftarrow \nu_{t-\tau}^l (\theta_{t-\tau}) + \frac{\partial \nu_{t-\tau}^l (\theta_{t-\tau})}{\partial \theta_{t-\tau}^l } (\theta_{t}^l -\theta_{t-\tau}^l)$ \hfill //approximation from recent iterations \\
}

$\bar{\mu}^l_{t,k}(\theta_t) \leftarrow \frac{1}{k} \sum_{\tau=0}^{k-1} \mu_{t-\tau}^l(\theta_t)$ \hfill //averaging over recent iterations\\

$\bar{\nu}^l_{t,k}(\theta_t) \leftarrow \frac{1}{k} \sum_{\tau=0}^{k-1} \max\big[ \nu_{t-\tau}^l(\theta_t), \mu_{t-\tau}^l(\theta_t)^2 \big ]$ \hfill //validation and averaging over recent iterations\\

$\bar{\sigma}^l_{t,k}(\theta_t)^2 \leftarrow \bar{\nu}^l_{t,k}(\theta_t) - \bar{\mu}_{t,k}^l(\theta_t)^2$ \hfill  \\

$\hat{x}^l_{t,i}(\theta_t)=\frac{x^l_{t,i}(\theta_t)-\bar{\mu}^l_{t,k}(\theta_t)}{\sqrt{\bar{\sigma}^l_{t,k}(\theta_t)^2 + \epsilon}}$ \hfill //normalize\\
$y^l_{t,i}(\theta_t) \leftarrow \gamma \hat{x}^l_{t,i}(\theta_t) + \beta$ \hfill //scale and shift\\
\end{algorithm}

\section{Efficient Implementation of ${\partial \mu_{t-\tau}^l (\theta_{t-\tau})}/{\partial \theta_{t-\tau}^l }$ and ${\partial \nu_{t-\tau}^l (\theta_{t-\tau})}/{\partial \theta_{t-\tau}^l }$}\label{appendix:efficient-imp}

Let $C_{out}$ and $C_{in}$ denote the output and input channel dimension of the $l$-th layer, respectively, and $K$ denotes the kernel size of $\theta_{t-\tau}^l$. $\mu_{t-\tau}^l$ and $\nu_{t-\tau}^l$ are thus of $C_{out}$ dimensions in channels, and $\theta_{t-\tau}^l$ is a $C_{out} \times C_{in} \times K$ dimensional tensor. A naive implementation of ${\partial \mu_{t-\tau}^l (\theta_{t-\tau})}/{\partial \theta_{t-\tau}^l }$ and ${\partial \nu_{t-\tau}^l (\theta_{t-\tau})}/{\partial \theta_{t-\tau}^l }$ involves computational overhead of $O(C_{out}\times C_{out} \times C_{in} \times K)$. Here we find that the operations of $\mu$ and $\nu$ can be implemented efficiently in $O(C_{in} \times K)$ and $O(C_{out} \times C_{in} \times K)$, respectively, thanks to the averaging of feature responses in $\mu$ and $\nu$.

Here we derive the efficient implementation of ${\partial \mu_{t-\tau}^l (\theta_{t-\tau})}/{\partial \theta_{t-\tau}^l }$. That of ${\partial \nu_{t-\tau}^l (\theta_{t-\tau})}/{\partial \theta_{t-\tau}^l }$ is about the same. Let us first simplify the notations a bit. Let ${ \mu^l}$ and ${ \theta^l }$ denote ${\mu_{t-\tau}^l (\theta_{t-\tau})}$ and ${\theta_{t-\tau}^l }$ respectively, by removing the irrelevant notations for iterations. The element-wise computation in the forward pass can be computed as
\begin{equation}
    \mu_j^l = \frac{1}{m}\sum_{i=1}^{m} x_{i,j}^l,
\label{eq:efficient_mu}
\end{equation}
where $\mu_j^l$ denotes the $j$-th channel in $\mu^l$, and $x_{i,j}^l$ denotes the $j$-th channel in the $i$-th example. $x_{i,j}^l$ is computed as
\begin{equation}
    x_{i,j}^l = \sum_{n=1}^{C_{in}} \sum_{k=1}^{K} \theta^l_{j,n,k} \cdot y^{l-1}_{i+\text{offset}(k),n},
\label{eq:efficient_x}
\end{equation}
where $n$ and $k$ enumerate the input feature dimension and the convolution kernel index, respectively, $\text{offset}(k)$ denotes the spatial offset in applying the $k$-th kernel, and $y^{l-1}$ is the output of the $(l-1)$-th layer.

The element-wise calculation of ${\partial \mu^l }/{\partial \theta^l } \in\mathbb{R}^{C_{out}\times C_{out}\times C_{in}\times K}$ is as follows, taking Eq.~\eqref{eq:efficient_mu} and Eq.~\eqref{eq:efficient_x} into consideration:
\begin{align}
\begin{split}
    [\frac{\partial \mu^l }{\partial \theta^l }]_{j,q,p,\eta} & = \frac{\partial \mu_j^l}{\partial \theta_{q,p,\eta}^l} \\
    & = \frac{\partial\frac{1}{m}\sum_{i=1}^{m} x_{i,j}^l}{\partial \theta_{q,p,\eta}^l} \\
    & = \frac{\partial \frac{1}{m}\sum_{i=1}^{m} \sum_{n=1}^{C_{in}} \sum_{k=1}^{K} \theta^l_{j,n,k}\cdot y^{l-1}_{i+\text{offset}(k),n}}{\partial \theta_{q,p,\eta}^l} \\
    & = \left\{\begin{array}{lc}\frac1m\sum\nolimits_{i=1}^my_{i+\text{offset}(\eta),p}^{l-1}&,\;j=q\\0&,\;j\neq q\end{array}\right..
\end{split}
\end{align}
Thus, $[\frac{\partial \mu^l }{\partial \theta^l }]_{j,q,p,\eta}$ takes non-zero values only when $j=q$. This operation can be implemented efficiently in $O(C_{in} \times K)$.
Similarly, the calculation of ${\partial \nu^l }/{\partial \theta^l }$ can be obtained in $O(C_{out} \times C_{in} \times K)$.

\section{Observation of the gradients diminishing}
\label{sec:grad-exp}
The key assumption in Eq.~\eqref{eq:taylor_mu_layer} and Eq.~\eqref{eq:taylor_nu_layer} is that for a node at the $l$-th layer, the gradient of its statistics with respect to the network weights at the $l$-th layer is larger than that of weights from the prior layers, i.e., 
\begin{align*}
&||g_{\mu}(l|l,t,\tau)||_F \gg ||g_{\mu}(r|l,t,\tau)||_F\\ &||g_{\nu}(l|l,t,\tau)||_F \gg ||g_{\nu}(r|l,t,\tau)||_F,&r<l
\end{align*}
where  $g_{\mu}(r|l,t,\tau)$ denotes $\frac{{\partial \mu_{t-\tau}^l(\theta_{t-\tau})}}{\partial \theta^r_{t-\tau}}$, $g_{\nu}(r|l,t,\tau)$ denotes $\frac{{\partial \nu_{t-\tau}^l(\theta_{t-\tau})}}{\partial \theta^r_{t-\tau}}$, and $||\cdot||_F$ denotes the Frobenius norm.

\begin{figure}[t]
    \centering
    \subfigure[The gradients of $\mu$]{
        \includegraphics[width=0.47\columnwidth]{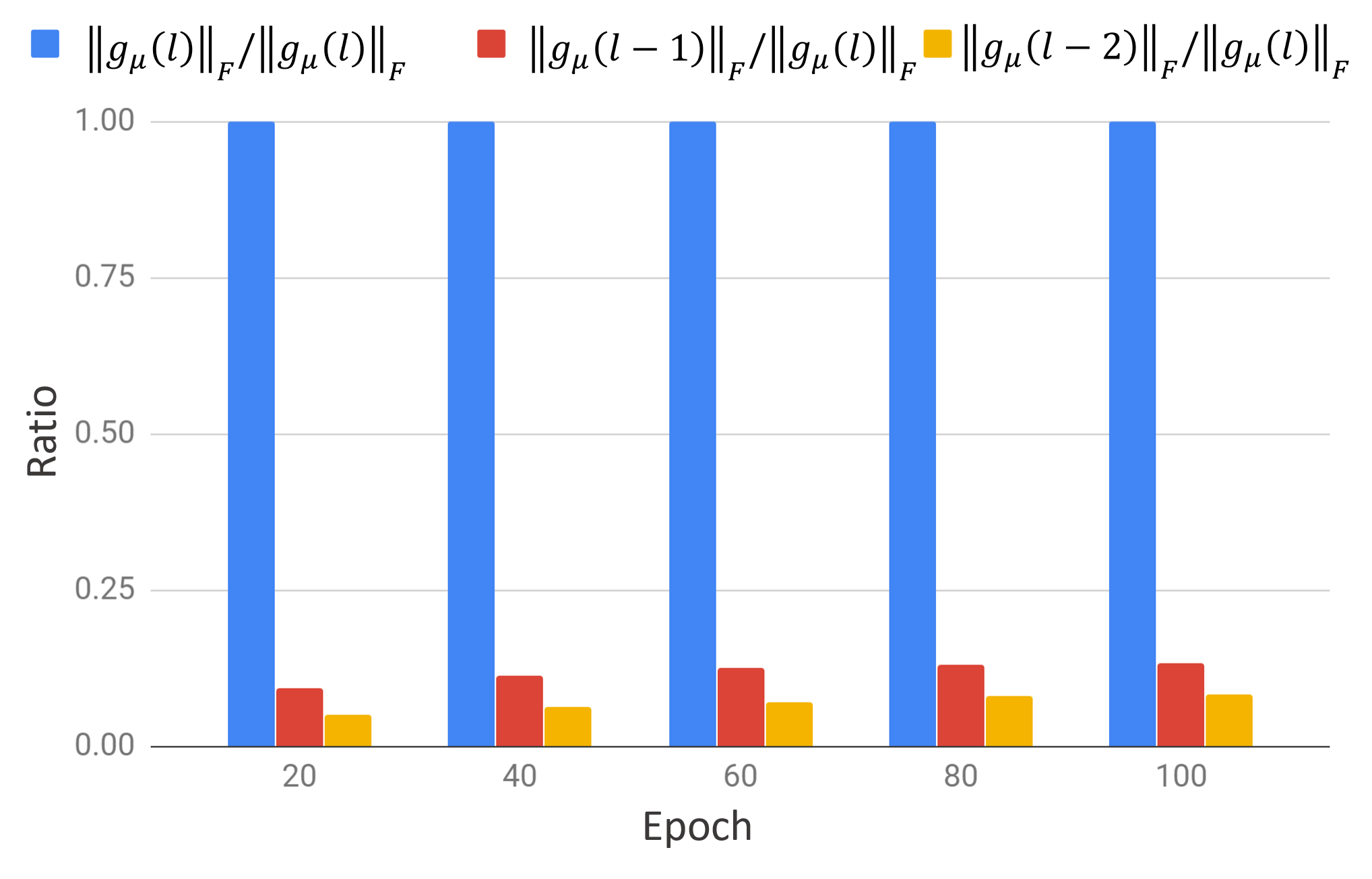}
        \label{fig:grad-mean-cifar}
    }\hfil
    \subfigure[The gradients of $\nu$]{
        \includegraphics[width=0.47\columnwidth]{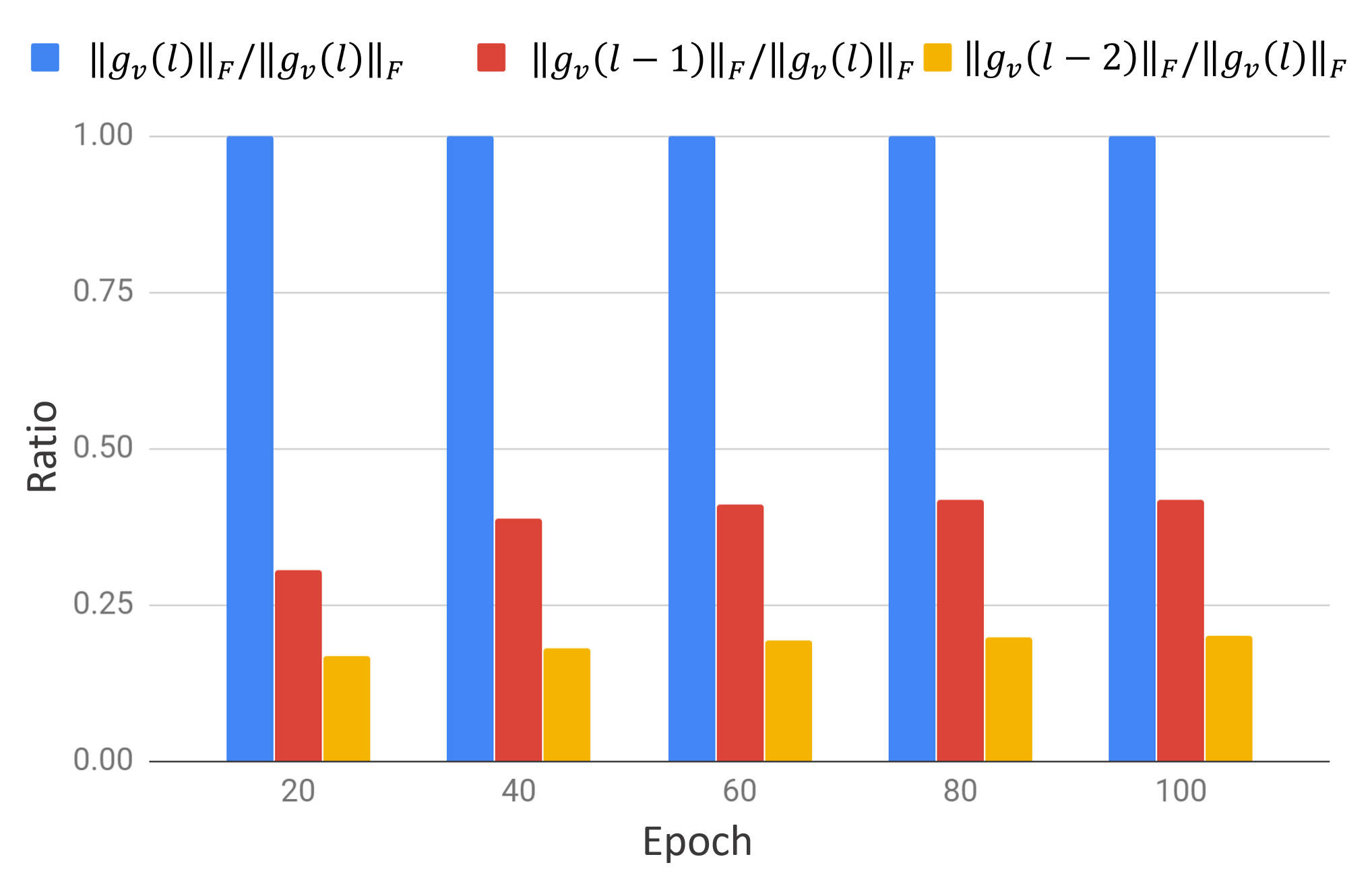}
        \label{fig:grad-mean-coco}
    }
    \caption{Comparison of gradients of statistics w.r.t. current layer vs. that w.r.t. previous layers on ImageNet.}
    \label{fig:grad-mu-nu}
\end{figure}

Here, we examine this assumption empirically for networks trained on ImageNet image recognition.
Both $||g_{\mu}(r)||_F / ||g_{\mu}(l)||_F $ and $||g_{\nu}(r)||_F / ||g_{\nu}(l)||_F $ for $r\in \{l-1, l-2\}$ are averaged over all CBN layers of the network at different training epochs (Figure \ref{fig:grad-mu-nu}). The results suggest that the key assumption holds well, thus validating the approximation in Eq.~\eqref{eq:taylor_mu_layer} and Eq.~\eqref{eq:taylor_nu_layer}.

We also study the gradients of non-ResNet models. The ratios of $||g_{\mu}||_F$ and $||g_{\nu}||_F$ are (0.20 and 0.41) for VGG-16 and (0.15 and 0.37) for Inception-V3, which is similar to ResNet (0.12 and 0.39), indicating that the assumption should also hold for the VGG and Inception series.

\end{document}